\documentclass[10pt,twocolumn,letterpaper]{article}

\usepackage[pagenumbers]{cvpr}

\usepackage{algorithm}
\usepackage{algorithmic}
\usepackage{amsfonts}
\usepackage{multirow}

\newcommand{\sref}[1]{§\ref{#1}}
\newcommand{\eref}[1]{Eq.~\ref{#1}}
\newcommand{\tref}[1]{Table~\ref{#1}}
\newcommand{\fref}[1]{Figure~\ref{#1}}

\definecolor{cvprblue}{rgb}{0.21,0.49,0.74}
\usepackage[pagebackref,breaklinks,colorlinks,allcolors=cvprblue]{hyperref}

\newif\ifwithsupp
\withsupptrue

\def\confName{CVPR}
\def\confYear{2027}

\title{Fast and Strong Image Protection against Diffusion Personalization}

\author{Jiwon Park \qquad Jihye Kang \qquad Namhyuk Ahn\\
Inha University}

\begin{document}
\maketitle

\begin{abstract}
Personalized text-to-image diffusion models can reproduce a person from a handful of photographs.
Image protection perturbs each image before publication so that fine-tuning on it fails, but it optimizes every perturbation from scratch, so its cost grows with the number of images.
We present the protective diffusion model (PDM), which is trained once and samples the perturbation of a new image over many denoising steps.
PDM distills an iterative teacher and is then post-trained with the target residual projection (TRP) loss as a reward, a stronger objective than existing ones.
PDM outperforms nine baselines on faces and artworks and nearly matches its teacher at a tenth of the cost.
\end{abstract}

\section{Introduction}

Personalized text-to-image diffusion models can reproduce a specific person from a handful of photographs, and they are now widely used~\cite{ruiz2023dreambooth,gal2022image,hu2021lora}.
A few public selfies are enough for these models to depict someone in a scene they never entered.
This capability underlies the non-consensual imagery and impersonation fraud that now affect ordinary individuals~\cite{umbach2024non,marchal2024generative}.
Because detection and takedown occur only after the images have circulated, the only timely defense is one applied before publication.

Image protection acts at that point, adding a perturbation within a fixed $L_\infty$ budget to every image before it is shared, so that a model fine-tuned on the protected copies fails to reproduce the subject~\cite{van2023anti,liu2024metacloak}.
Across these methods the perturbation is produced in the same way, by solving an iterative optimization over the pixels of each image from scratch~\cite{liang2023adversarial,liang2023mist,xue2023toward,zheng2023understanding}.
Because the optimization runs per image, its cost scales with the number of photographs.
Protecting a single image takes from several seconds to tens of minutes, so a subject protecting a photo album pays that price once for every photograph in it.

\begin{figure}
    \centering
    \includegraphics[width=\linewidth]{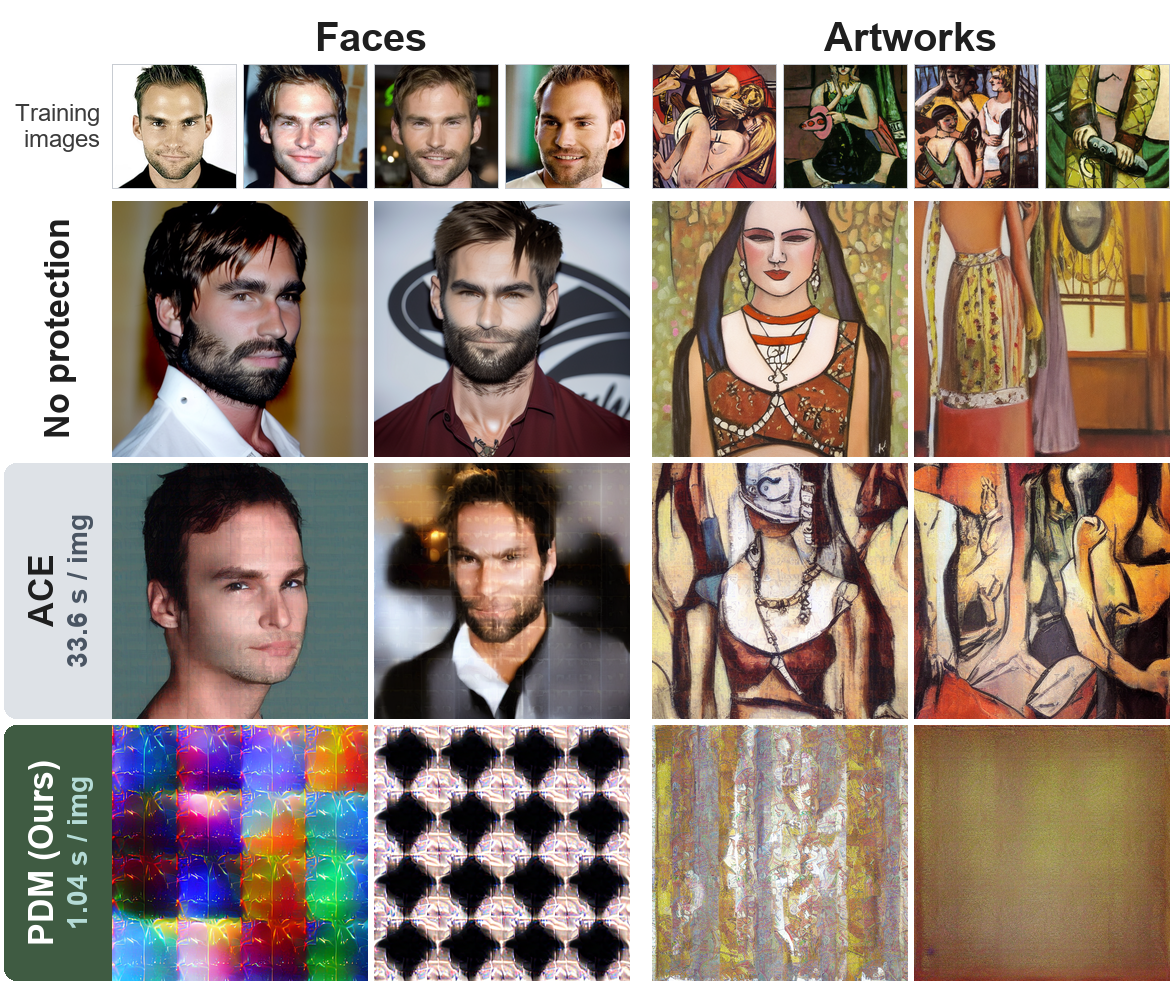}
    \caption{\textbf{PDM: Fast and strong image protection.} Generations of a model personalized on the protected images. PDM keeps the personalized model from reproducing the subject.}
    \vspace{-7pt}
    \label{fig:teaser}
\end{figure}

\begin{figure*}[t]
\centering
\includegraphics[width=\textwidth]{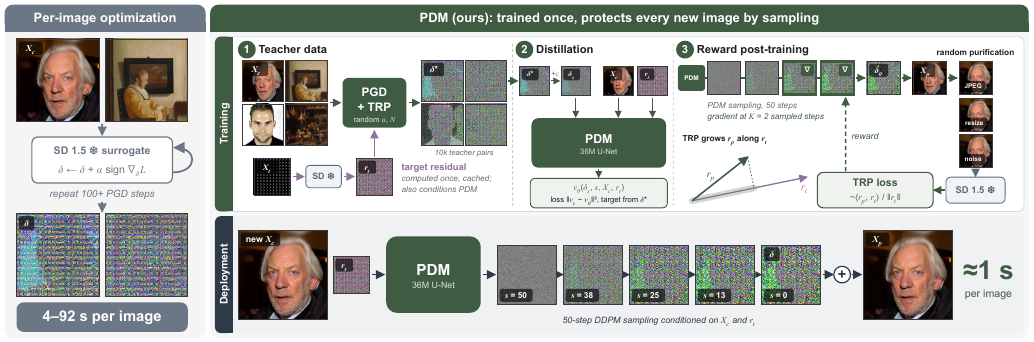}
\caption{Method overview. \textbf{(Left)} Per-image optimization runs a separate PGD search through the frozen surrogate for every image. \textbf{(Right)} PDM is trained once. A PGD teacher under the TRP loss produces perturbations with a randomized step size and step count (1), the generator is distilled on them (2), and it is then post-trained with the TRP loss as a reward, evaluated after a random purification (3). At deployment PDM protects a new image by 50-step sampling, in about one second.}
\vspace{-2pt}
\label{fig:overview}
\end{figure*}

A natural response is to amortize the cost by training a model once, so that protecting a new image requires no per-image optimization.
How much protection such a model achieves depends on two factors, the capacity of the generator and the strength of its training signal.
An iterative optimizer refines its perturbation over many steps.
Existing amortized protectors produce their perturbation in one step, either from fixed pretrained perturbations or by a single forward pass~\cite{guo2024real,ahn2025nearly}, and a one-step generator stays short of the iterative optimizer it learns from (\sref{sec:analysis}).
A generator trained to imitate a teacher is also bounded by that teacher, so the protective objective that defines the teacher sets a ceiling on the generator.
A strong amortized protector therefore needs a generator that refines its perturbation over many steps and an objective strong enough to make a good teacher and to optimize directly once imitation ends.

We address both challenges with the proposed \textbf{protective diffusion model (PDM)}, a conditional diffusion model that is trained once and protects a new image by sampling its perturbation without per-image optimization.
Its denoising process refines the perturbation over many steps, as the iterative teacher does.
Training then proceeds in two stages.
First, we distill an iterative optimization teacher~\cite{madry2017towards}, which transfers the structure of the perturbations that optimization produces but bounds the model by the teacher it imitates.
We then post-train the distilled checkpoint using the protective objective as a differentiable reward, which lifts protection from the distilled model to close to the teacher's level at a tenth of its computational cost for every new image.

Both stages use the proposed \textbf{target residual projection (TRP) loss}, which acts on the \emph{denoising residual}, the factor of the personalization gradient that does not depend on how the attacker parameterizes the update.
TRP grows that residual along a target direction within the perturbation budget.
We implement the signal as a projection onto the target residual, which rewards every gain of the residual along the target direction.
A squared error toward the same target also rewards a smaller residual, which lowers the loss that the attacker minimizes.
Empirically, the squared error converts additional budget into little additional protection, while the projection converts it monotonically.
TRP protects more strongly than existing objectives, so it gives PDM a high-quality iterative teacher to distill and then becomes its reward.
Our contributions are as follows:

\begin{itemize}
\item We introduce PDM, to our knowledge the \textbf{first protector that samples its perturbation from a diffusion model}, trained by distillation and reward post-training.
\item We propose the \textbf{TRP loss}, which increases the denoising residual along a defender-chosen direction and is both the teacher's objective and the reward.
\item PDM \textbf{outperforms nine baselines} on faces and artworks and nearly matches its teacher at a tenth of the cost.
\end{itemize}

\section{Related Work}

\noindent\textbf{Protection against personalization.}
A protection adds a budget-bounded perturbation to an image so that a model fine-tuned on it fails to reproduce the subject's identity or style~\cite{ahn2024imperceptible,van2023anti,liang2023adversarial,liang2023mist,zheng2023understanding,salman2023raising,xue2023toward,shan2023glaze,liu2024metacloak,xu2024perturbing}.
These methods differ mainly in the objective they optimize, and our TRP loss is a new objective of this family.
Each of them optimizes the perturbation from scratch for every image, so the cost grows with the number of images.
Amortized protectors remove this cost by training a network once and applying it to each new image.
RID~\cite{guo2024real} trains a perturbation network against an ensemble of surrogates, and FastProtect~\cite{ahn2025nearly} selects among a mixture of pretrained perturbations by the latent code of the input.
Both produce the perturbation in one step.
PDM is also trained once, but it generates the perturbation with a diffusion model, which gives the generator more capacity to reproduce the perturbations of iterative optimization.

An attacker can also purify the published images before personalization, with light transforms such as JPEG compression or with an off-the-shelf generative model~\cite{pleimling2026off,honig2025adversarial}.
Per-image protections resist purification by optimizing under random transformations~\cite{liu2024metacloak}, which adds to the cost of every image.
PDM applies a random purification inside its training reward, so it learns perturbations robust to light purification at no cost at deployment.

\smallskip
\noindent\textbf{Reward-based diffusion training.}
We train the generator by distilling a teacher and then directly optimizing the protective objective.
This approach draws on work that fine-tunes diffusion models using a differentiable reward~\cite{clark2024directly} and on score distillation, in which a pretrained diffusion model supplies the training signal for another model~\cite{poole2022dreamfusion}.
Both lines of work target generation quality, and we adapt their machinery to supervise a protector by treating the protective objective itself as the reward.
\section{Preliminaries}

\noindent\textbf{Diffusion personalization and protection.}
We work with a latent diffusion model~\cite{rombach2022high} that encodes an image $X$ to $z = \mathcal{E}(X)$ and trains a noise predictor $\epsilon_\theta$ on the forward process $z_\tau = \alpha_\tau z + \sigma_\tau \epsilon$ with $\epsilon \sim \mathcal{N}(0, I)$.
The training objective on a single sample is $\mathcal{L}_{\mathrm{ft}}(X) = \left\| \epsilon_\theta(z_\tau, \tau, c) - \epsilon \right\|_2^2$.
Personalization adapts $\epsilon_\theta$ to a subject from a handful of images~\cite{ruiz2023dreambooth,gal2022image}, usually by training a LoRA adapter~\cite{hu2021lora}.

The defender adds a perturbation $\delta$ to a clean image $X_c$ under a budget $\|\delta\|_\infty \le \eta$ and publishes the protected image $X_p = X_c + \delta$.
The attacker collects the published set $\mathcal{S}_p$ and may purify it with a transform that does not depend on the protection method.
The attacker then fine-tunes the diffusion model to $\theta^\star = \arg\min_\theta \mathbb{E}_{X \in \mathcal{S}_p}[\mathcal{L}_{\mathrm{ft}}(X; \theta)]$, and protection succeeds when $\theta^\star$ reproduces the subject poorly.

\smallskip
\noindent\textbf{Denoising residual.}
For an image $X$, the surrogate prediction minus the sampled noise, $r_X = \hat\epsilon_X - \epsilon$, is its \emph{denoising residual}.
Since $\nabla_{\hat\epsilon_X} \mathcal{L}_{\mathrm{ft}}(X) = 2 r_X$, the parameter gradient of the personalization loss factors as $\nabla_\theta \mathcal{L}_{\mathrm{ft}}(X) = 2 J_\theta(X)^\top r_X$, where $J_\theta(X)$ is the Jacobian of the prediction with respect to the parameters.
Subscripts tie these quantities to images, so that $r_p$, $\hat\epsilon_p$, and $\epsilon_p$ denote the residual, the prediction, and the sampled noise of a protected image $X_p$.

\section{Method}
PDM is a conditional diffusion model that samples a perturbation $\delta$ for a clean image $X_c$ (\fref{fig:overview}).
We first propose the TRP loss as its protective objective (\sref{sec:trp}) and then train PDM with this loss in two stages (\sref{sec:pgm}).

\subsection{Target Residual Projection Loss}
\label{sec:trp}

\noindent\textbf{Protective objectives in residual space.}
We build the protective objective on the residual factor $r_p$ of the personalization gradient $2 J_\theta(X_p)^\top r_p$.
At the pretrained weights, the residual is the same whichever parameters the attacker adapts, whereas the Jacobian depends on this choice.
The objective therefore evaluates $r_p$ with the frozen surrogate at the pretrained weights and is a proxy for the parameter-level signal.
Prior work explains that a model fine-tuned on protected images learns a bias that reverses their error pattern and reproduces this bias in its generations~\cite{zheng2023understanding}.
In this account, a common target gives the errors of all protected images one shared pattern and hence one consistent bias, whereas untargeted errors differ from image to image and do not form one bias.
Following this account, we push the residual of every protected image along one fixed target direction.
We write a protective objective as a comparison $\mathcal{D}(r_p, q)$ between the residual $r_p$ and a reference $q$, so TRP is defined by two choices, the reference that sets this direction and the geometry $\mathcal{D}$ that compares the residual with it.

\smallskip
\noindent\textbf{Target reference.}
The defender chooses a target image $X_t$, and TRP takes the denoising residual of this image as the reference.
ACE~\cite{zheng2023understanding} matches the prediction of the protected image to the latent $\mathcal{T}$ of a target image through the squared error ${\|\hat\epsilon_p - \mathcal{T}\|_2^2}$, and substituting $\hat\epsilon_p = r_p + \epsilon_p$ turns this objective into the squared distance between $r_p$ and $\mathcal{T} - \epsilon_p$.
The two references are therefore
\begin{equation}
q_{\mathrm{TRP}} = r_t = \hat\epsilon_t - \epsilon_t,
\qquad
q_{\mathrm{ACE}} = \mathcal{T} - \epsilon_p .
\label{eq:refs}
\end{equation}
Both references fix one direction for every protected image and differ in where this direction lies.
We compute $r_t$ once with the frozen surrogate and average it over timesteps and noise samples, so every protected image uses the same $r_t$ (details in Suppl.).
The ACE reference averages to the fixed latent $\mathcal{T}$ over the noise draws, so its direction is the latent of the target image.
TRP replaces this direction with the residual of the target image, which lies in the same space as $r_p$, and we call this choice \emph{direct residual targeting}.

\smallskip
\noindent\textbf{Geometry.}
Given $q_{\mathrm{TRP}} = r_t$, the remaining choice is the geometry $\mathcal{D}$.
With $r_t$ as the reference, the squared error that ACE uses expands as
\begin{equation}
\underbrace{\left\|r_p - r_t\right\|_2^2}_{\text{exact matching}} = \underbrace{\left\|r_p\right\|_2^2}_{\text{magnitude}} - 2\underbrace{\left\langle r_p, r_t \right\rangle}_{\text{alignment}} + \left\|r_t\right\|_2^2 .
\label{eq:mse}
\end{equation}

These terms separate the design space.
Untargeted protections~\cite{liang2023adversarial,van2023anti} choose $\delta$ to raise the magnitude term alone, and ACE chooses $\delta$ to lower the whole expression with the reference $q_{\mathrm{ACE}}$ of \eref{eq:refs}.
Exact matching with $q_{\mathrm{TRP}}$ lowers the same expression.
The alignment term rewards a perturbation that grows $r_p$ along $r_t$, which is the push that this account calls for.
The magnitude term penalizes the size of $r_p$ in every direction.
This term is the personalization loss $\mathcal{L}_{\mathrm{ft}}$ that the attacker minimizes, so exact matching also rewards a perturbation that lowers this loss.

Viewed in residual space, exact matching moves $r_p$ toward $r_t$, that is, along the difference $r_t - r_p$, which has two parts.
The part $r_t$ pushes the residual along the target, and the part $-r_p$ shrinks the current residual, which lowers the loss of the attacker at the pretrained weights (\tref{tab:budget}).
Scaling up the target in the objective strengthens the first part relative to the second, and in the limit only the push along $r_t$ remains.
This limit is the projection geometry, which keeps only the push along the target.

\smallskip
\noindent\textbf{TRP loss.}
Taking the direct target reference and the projection geometry gives the proposed TRP loss,
\begin{equation}
\mathcal{L}_{\mathrm{TRP}} = -\,\frac{\langle r_p, r_t\rangle}{\|r_t\|_2},
\quad \|\delta\|_\infty \le \eta,
\label{eq:trp}
\end{equation}
averaged over surrogate timesteps and noise samples.
The negative of the loss is the scalar projection of $r_p$ onto the target direction and equals $\|r_p\|_2 \cos\angle(r_p, r_t)$.
It grows with the component of $r_p$ along $r_t$ and does not depend on the other components, so the loss rewards every gain along the target direction and has no term that shrinks the residual.
Dividing by $\|r_t\|_2$ removes the dependence on the scale of the target residual, so the objective depends on the target only through its direction.
The objectives thus differ in how they move the residual.
Untargeted protections raise the attacker's loss $\mathcal{L}_{\mathrm{ft}}$ in no fixed direction, exact matching and ACE add a squared magnitude term to their push toward the target, and TRP keeps only the push along $r_t$.

\subsection{Protective Diffusion Model}
\label{sec:pgm}

An iterative optimizer evaluates the objective on the protected image at every step, whereas PDM samples its perturbation without evaluating any objective at deployment.
The objective must therefore enter PDM during training.
We bring it in through two stages, first by distilling the perturbations that optimization produces and then by optimizing the TRP loss on the generator's own samples, with a random purification inside the reward.

\smallskip
\noindent\textbf{Generator.}
PDM is a 36M U-Net~\cite{ronneberger2015u} trained from scratch with the $v$-prediction parameterization, independent of the surrogate.
It runs its own diffusion process over perturbations, and we write $s$ for its steps and keep $\tau$ for the surrogate timestep.
The clean image $X_c$ enters by concatenation with the noised perturbation $\delta_s$, and the target residual $r_t$ is injected at the $64\times64$ resolution, so one generator protects any image.
At deployment PDM draws a 50-step sample, which is a complete protection on its own.

\smallskip
\noindent\textbf{Distillation pre-training.}
Reward training from a random initialization is unstable and reaches much weaker protection (\tref{tab:ablation}), so we first pretrain the generator by distillation.
The teacher data are generated by solving
\begin{equation}
\delta^\star = \arg\min_{\|\delta\|_\infty \le \eta} \mathcal{L}_{\mathrm{TRP}}\!\left(r_p(\delta), r_t\right)
\label{eq:teacher}
\end{equation}
with PGD~\cite{madry2017towards}, and the resulting triples $(X_c, r_t, \delta^\star)$ form the training set.
We randomize the step size and the number of steps across examples, since teacher data drawn under a single fixed configuration concentrates the solutions into a nearly deterministic mapping, leaving the generator no distribution to learn.
The generator is then fit to these perturbations, treating $\delta^\star$ as the data and $(X_c, r_t)$ as conditioning,
\begin{equation}
\mathcal{L}_{\mathrm{distil}} = \mathbb{E}_{s, \epsilon}\!\left[\left\|v_s - v_\theta(\delta_s, s, X_c, r_t)\right\|_2^2\right],
\label{eq:distil}
\end{equation}
where $v_s = \alpha_s \epsilon - \sigma_s \delta^\star$ is the velocity target at sampling step $s$.
The distilled generator already samples perturbations whose structure matches what the teacher produces on the same image.
However, the generator only imitates the perturbations that the teacher outputs and never optimizes the protective objective directly, so the distilled generator is suboptimal under that objective, and its protection stays well below the teacher's.

\smallskip
\noindent\textbf{Reward post-training.}
We then optimize the TRP loss directly on the generator's own samples, using it as a differentiable reward as in reward fine-tuning of diffusion models~\cite{clark2024directly}.
Backpropagating through the full 50-step sampling chain exceeds the available memory, so we take the gradient through single denoising steps, inspired by ReFL~\cite{xu2023imagereward}.
Each iteration runs the sampler without gradients and picks $K = 2$ of its steps at random, excluding the first $30\%$ of the trajectory, where the perturbation is still dominated by noise.
At a picked step $s_k$, the network predicts the velocity with gradients, and the scheduler converts it into an estimate $\hat\delta_0$ of the clean perturbation. The reward is evaluated on the protected image $\hat X_p = X_c + \Pi_\eta(\hat\delta_0)$,
\begin{equation}
\mathcal{L}_{\mathrm{post}} = \mathbb{E}_{s_k,\, \phi \sim \mathcal{P}}\!\left[\mathcal{L}_{\mathrm{TRP}}\!\left(r_p(\phi(\hat X_p)), r_t\right)\right],
\label{eq:post}
\end{equation}
where $\Pi_\eta$ projects onto the budget and $\phi$ is a purification drawn from a set $\mathcal{P}$.
The surrogate evaluates $r_p$ at a random timestep and noise sample, as in the teacher's PGD, so $r_p$ and $r_t$ share the surrogate's noise space.

\smallskip
\noindent\textbf{Purification-aware reward.}
An attacker may purify the published images before fine-tuning, and a perturbation optimized on the unpurified image loses much of its protection even under a simple transformation such as JPEG compression~\cite{liu2024metacloak,zhu2025gapdiff}.
Per-image methods resist purification by optimizing under random transformations~\cite{liu2024metacloak}, which adds to the cost of every image.
PDM moves this cost into training.
The set $\mathcal{P}$ in \eref{eq:post} models the purifications that an attacker can apply at almost no cost, namely JPEG compression, resizing, and additive noise, together with the identity (no purification).
Each reward evaluation thus simulates an attacker who purifies the protected image first, so the generator gains reward only for protection that survives purification.
The teacher never sees a purification, so this robustness comes from the reward stage alone, and it adds no cost at deployment because sampling is unchanged.

\section{Experiments}

\subsection{Experimental Setup}

\begin{table*}[t]
\centering
\caption{\textbf{Protection method comparison on both domains.} FSS = (1 $-$ FDFR) $\times$ ISM, and CSD is the style similarity to the artist's images. In each domain, PSNR and LPIPS compare each protected image with its clean copy (imperceptibility), and the other metrics measure the protection after personalization. Time and peak memory are for protecting one $512\times512$ image on one RTX 5090 without model loading, and the time of ACE, Anti-DB, CAAT, and MetaCloak includes their per-subject preparation. $^\ast$Amortized protector, released weights trained on faces only. $^\ddagger$PDM with reduced perturbation strength, matching the visibility of the baselines (\fref{fig:impercept}). Bold and underline mark the best and second-best protections.}
\label{tab:comparison}
\footnotesize
\setlength{\tabcolsep}{1.8pt}
\begin{tabular*}{\textwidth}{@{\extracolsep{\fill}}lcccccccccccccc@{}}
\toprule
\multirow{2}{*}{Method} & \multicolumn{2}{c}{Cost} & \multicolumn{7}{c}{CelebA-HQ (faces)} & \multicolumn{5}{c}{WikiArt (artworks)} \\
\cmidrule(lr){2-3}\cmidrule(lr){4-10}\cmidrule(l){11-15}
& Time (s) $\downarrow$ & Mem.\ (GB) $\downarrow$ & PSNR $\uparrow$ & LPIPS $\downarrow$ & FSS $\downarrow$ & FDFR $\uparrow$ & ISM $\downarrow$ & FID $\uparrow$ & BRISQUE $\uparrow$ & PSNR $\uparrow$ & LPIPS $\downarrow$ & CSD $\downarrow$ & CLIP-SIM $\downarrow$ & FID $\uparrow$ \\
\midrule
No protection & -- & -- & -- & -- & 0.443 & 0.025 & 0.454 & 80.6 & 14.4 & -- & -- & 0.379 & 0.774 & 136.9 \\
\midrule
PhotoGuard~\cite{salman2023raising} & 4.27 & 4.88 & 34.59 & 0.226 & 0.151 & 0.159 & 0.175 & 176.7 & \underline{31.5} & 34.26 & 0.115 & 0.254 & 0.693 & 192.3 \\
Mist~\cite{liang2023mist} & 11.25 & 5.11 & 34.71 & 0.224 & 0.152 & 0.134 & 0.177 & 178.6 & 30.1 & 34.39 & 0.114 & 0.263 & 0.695 & 194.4 \\
SDS ($-$)~\cite{xue2023toward} & 5.46 & 4.56 & \textbf{35.55} & \textbf{0.174} & 0.498 & 0.006 & 0.502 & 75.5 & 22.8 & \underline{35.04} & \textbf{0.094} & 0.293 & 0.737 & 155.8 \\
AdvDM~\cite{liang2023adversarial} & 11.10 & 5.11 & 35.16 & 0.219 & 0.345 & 0.003 & 0.346 & 110.8 & 15.6 & 34.79 & 0.137 & 0.359 & 0.781 & 139.0 \\
ACE~\cite{zheng2023understanding} & 33.63 & 7.08 & \underline{35.50} & 0.198 & 0.185 & 0.138 & 0.214 & 167.0 & 21.8 & \textbf{35.09} & \underline{0.107} & 0.329 & 0.741 & 176.1 \\
Anti-DB~\cite{van2023anti} & 92.36 & 14.57 & 34.83 & 0.221 & 0.311 & 0.050 & 0.325 & 130.1 & 10.0 & 34.40 & 0.144 & 0.324 & 0.767 & 144.7 \\
CAAT~\cite{xu2024perturbing} & 29.91 & 18.58 & 34.88 & 0.221 & 0.346 & 0.034 & 0.358 & 111.2 & 25.8 & 34.49 & 0.142 & 0.351 & 0.779 & 142.8 \\
MetaCloak~\cite{liu2024metacloak} & 1666.56 & 16.50 & 34.56 & 0.240 & 0.331 & 0.034 & 0.343 & 122.3 & 12.0 & 34.34 & 0.165 & 0.330 & 0.771 & 140.1 \\
RID$^\ast$~\cite{guo2024real} & \textbf{0.26} & \textbf{2.10} & 34.91 & 0.219 & 0.362 & 0.050 & 0.380 & 95.3 & 23.0 & 34.36 & 0.164 & 0.374 & 0.779 & 142.7 \\
\midrule
\textbf{PDM} & \underline{1.04} & \underline{2.80} & 33.56 & 0.244 & \textbf{0.061} & \textbf{0.394} & \textbf{0.106} & \textbf{212.4} & \textbf{32.4} & 33.21 & 0.164 & \textbf{0.156} & \textbf{0.658} & \textbf{242.3} \\
\textbf{PDM$^\ddagger$} & \underline{1.04} & \underline{2.80} & 35.13 & \underline{0.191} & \underline{0.084} & \underline{0.266} & \underline{0.119} & \underline{183.5} & 29.2 & 34.79 & 0.123 & \underline{0.207} & \underline{0.681} & \underline{204.4} \\
\bottomrule
\end{tabular*}
\end{table*}

\noindent\textbf{Datasets.}
We evaluate on two domains, faces from CelebA-HQ~\cite{karras2018progressive} and artworks from WikiArt~\cite{saleh2015large}, following the protocols of~\cite{ahn2024imperceptible}.
Each domain has 20 subjects with 4 images each, where a subject is one identity for faces and one artist for artworks.
The main comparison with the baselines and the sampling-step sweep use all 20 subjects, and the purification and transfer studies and the model analysis use the first 10 subjects of each domain.
For each subject we add a protective perturbation to every image before publication and personalize a model on the protected copies.

\smallskip
\noindent\textbf{Baselines.}
All methods protect each image under the budget $\eta = 12/255$.
We compare against eight per-image optimization baselines, PhotoGuard~\cite{salman2023raising}, Mist~\cite{liang2023mist}, SDS~\cite{xue2023toward}, AdvDM~\cite{liang2023adversarial}, ACE~\cite{zheng2023understanding}, Anti-DreamBooth (Anti-DB)~\cite{van2023anti}, CAAT~\cite{xu2024perturbing}, and MetaCloak~\cite{liu2024metacloak}.
We run each of them based on its official implementation and change only the budget and the surrogate to match PDM.
We also compare against RID~\cite{guo2024real}, an amortized protector, with its released weights.
We follow the setting that all baselines share, UNet latent diffusion models~\cite{rombach2022high}, and evaluate on SD 1.5 with transfer to SD 2.1~\cite{rombach2022high} and SDXL~\cite{podell2023sdxl}.
Details of each baseline are in Suppl.

\begin{figure}[t]
\centering
\includegraphics[width=\linewidth]{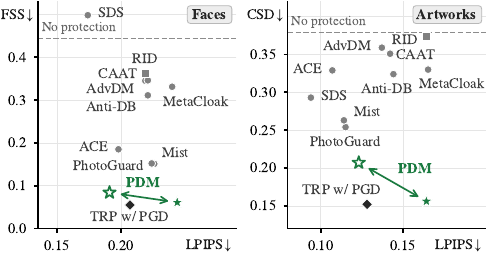}
\caption{\textbf{Protection against visibility.} LPIPS between each protected image and its clean copy (left is less visible) and the protection after personalization (lower is better). The arrow joins PDM (filled star) and the same model with reduced perturbation strength (hollow star, PDM$^\ddagger$ in \tref{tab:comparison}).}
\label{fig:impercept}
\end{figure}

\smallskip
\noindent\textbf{Evaluation.}
The victim fine-tunes SD 1.5 with LoRA~\cite{hu2021lora} on the protected images of each subject, with the prompt ``a photo of sks person'' for faces and ``a painting of sks style'' for artworks (full recipe in Suppl.).
We measure protection on the generations of the personalized model.
On faces we report FDFR ($\uparrow$), ISM ($\downarrow$), and their combination FSS~$= (1-\mathrm{FDFR})\times\mathrm{ISM}$ ($\downarrow$), computed per subject and then averaged.
On artworks we report CSD ($\downarrow$)~\cite{somepalli2024measuring} and CLIP-SIM ($\downarrow$)~\cite{radford2021learning}.
We also report BRISQUE ($\uparrow$)~\cite{mittal2012no} on faces and FID ($\uparrow$)~\cite{heusel2017gans} on both.
We measure imperceptibility with PSNR ($\uparrow$) and LPIPS ($\downarrow$)~\cite{zhang2018unreasonable}, computed between each protected image and its clean copy.

\begin{figure}[t]
\centering
\includegraphics[width=\linewidth]{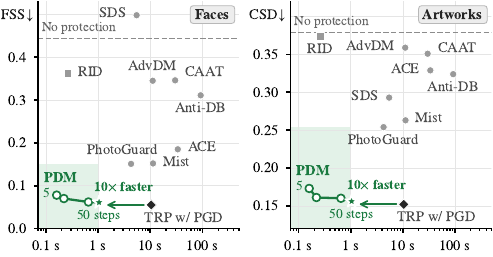}
\caption{\textbf{Protection against speed.} Time to protect one image (log scale, MetaCloak omitted) and the protection after personalization (lower is better). The PDM curve varies the number of DDPM sampling steps from 5 to 50. In the shaded region, protection takes under one second and is stronger than every baseline.}
\label{fig:steps}
\end{figure}

\smallskip
\noindent\textbf{Implementation details.}
All protections and evaluations run at $512\times512$ with SD 1.5~\cite{rombach2022high} as the frozen surrogate.
PDM protects an image with one 50-step sample, and all timings are measured on a single NVIDIA RTX 5090.
PDM and its PGD teacher use a single fixed target image, chosen once and kept for training and deployment.
The PGD teacher in our comparisons runs 100 steps of size $1/255$, inside the ranges randomized for the distillation data.
We will release the code and trained models. Detailed setups for training and evaluation are in Suppl.

\subsection{Model Comparison}
\label{sec:comparison}

\begin{figure*}[t]
\centering
\includegraphics[width=\textwidth]{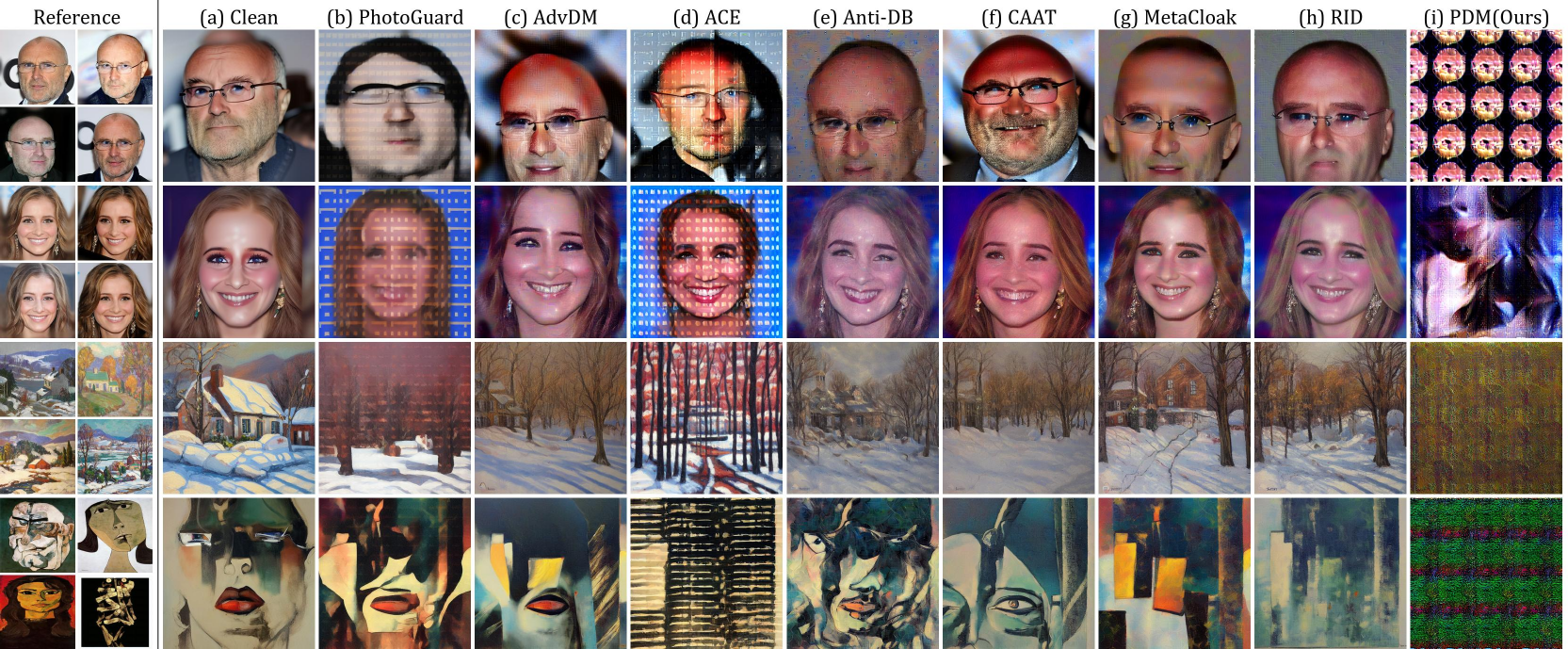}
\caption{\textbf{Qualitative comparison on faces and artworks.} The reference column shows the training images of each subject (top two rows: faces; bottom two rows: artworks), and the remaining columns show personalized generations under the corresponding protection, all from the same sampling seed within a row.}
\label{fig:qual}
\end{figure*}

\noindent\textbf{Quantitative comparison.}
In \tref{tab:comparison}, we compare the proposed PDM with the nine baselines on faces and artworks.
PDM runs no optimization on the image it protects, yet it gives the strongest protection on every protection metric of both domains, ahead of the eight baselines that optimize each image.
On faces, its FSS is less than half that of the strongest baseline.
Two parts of PDM explain this result.
The TRP loss supplies a teacher signal that persists through the fine-tuning of the victim, and the two training stages amortize this signal into one generator that reproduces it in a single sampling pass.
Strong protection thus no longer requires optimizing every image.

\smallskip
\noindent\textbf{Imperceptibility.}
A stronger protective perturbation protects more but is easier to see, so we compare protection together with imperceptibility.
We trace this trade-off for PDM by also scaling down its perturbation (PDM$^\ddagger$ in \tref{tab:comparison}), and \fref{fig:impercept} places each method by its LPIPS against its protection.
In both domains the two PDM points lie below every baseline, so PDM protects more than the baselines at either level of visibility.
Only its teacher, TRP optimized with PGD, lies slightly ahead of PDM, and this small gap is the cost of amortizing a 100-step optimization into a cheaper diffusion process.
By LPIPS, the scaled PDM is the second least visible method on faces and less visible than most baselines on artworks, while its protection stays far ahead of every baseline.
The margin of PDM in \tref{tab:comparison} therefore does not come from a more visible perturbation.

\begin{table*}[t]
\centering
\caption{\textbf{Robustness to purification.} FSS / CSD (lower is better) after the attacker purifies the published images before personalization. The upper block holds light purifications and the lower block generative purifications, and $^\dagger$ marks purifications not used when training PDM. PDM w/o PAR is the same generator post-trained for the same 5 epochs without the purification-aware reward (PAR), which scores each protected image after a random purification. Bold marks the best protection per row and metric. Mist is omitted because its encoder objective and its scores in \tref{tab:comparison} closely follow PhotoGuard.}
\label{tab:purify}
\footnotesize
\setlength{\tabcolsep}{2pt}
\begin{tabular*}{\textwidth}{@{\extracolsep{\fill}}lcccccccc@{}}
\toprule
Purification & No protection & PhotoGuard & ACE & Anti-DB & CAAT & MetaCloak & PDM w/o PAR & PDM \\
\midrule
None & 0.406 / 0.414 & 0.172 / 0.252 & 0.224 / 0.323 & 0.308 / 0.341 & 0.334 / 0.372 & 0.321 / 0.348 & 0.100 / 0.173 & \textbf{0.070} / \textbf{0.145} \\
\midrule
JPEG ($q = 75$) & 0.420 / 0.398 & 0.408 / 0.445 & 0.404 / 0.461 & 0.399 / 0.398 & 0.387 / 0.418 & 0.338 / 0.393 & 0.209 / 0.363 & \textbf{0.123} / \textbf{0.280} \\
Resize ($0.5\times$, bilinear) & 0.415 / 0.375 & 0.404 / 0.400 & 0.386 / 0.395 & 0.366 / 0.367 & 0.418 / 0.387 & 0.339 / 0.315 & 0.292 / 0.344 & \textbf{0.176} / \textbf{0.278} \\
WebP ($q = 75$)$^\dagger$ & 0.437 / 0.398 & 0.453 / 0.414 & 0.440 / 0.457 & 0.420 / 0.409 & 0.434 / 0.395 & 0.404 / 0.378 & 0.362 / 0.348 & \textbf{0.221} / \textbf{0.281} \\
Gaussian blur ($\sigma = 1$)$^\dagger$ & 0.400 / 0.387 & 0.416 / 0.403 & 0.365 / 0.411 & 0.379 / 0.356 & 0.415 / 0.367 & 0.323 / 0.323 & 0.240 / 0.334 & \textbf{0.165} / \textbf{0.260} \\
Resize ($0.5\times$, Lanczos)$^\dagger$ & 0.400 / 0.404 & 0.393 / 0.385 & 0.345 / 0.411 & 0.372 / 0.387 & 0.397 / 0.403 & 0.315 / 0.344 & 0.204 / 0.325 & \textbf{0.141} / \textbf{0.241} \\
\midrule
Noisy upscaling~\cite{honig2025adversarial}$^\dagger$ & 0.408 / 0.402 & 0.413 / 0.395 & \textbf{0.404} / 0.404 & 0.409 / \textbf{0.390} & 0.410 / 0.405 & 0.413 / 0.399 & 0.413 / 0.410 & 0.411 / 0.400 \\
FLUX img2img~\cite{pleimling2026off}$^\dagger$ & 0.389 / 0.362 & 0.410 / 0.356 & 0.406 / \textbf{0.336} & 0.430 / 0.377 & 0.417 / 0.373 & 0.376 / 0.368 & 0.406 / 0.381 & \textbf{0.373} / 0.368 \\
\bottomrule
\end{tabular*}
\end{table*}

\smallskip
\noindent\textbf{Efficiency.}
The first two columns of \tref{tab:comparison} give the cost of protecting one image, excluding model loading.
PDM samples a perturbation in 1.04\,s, four to five times faster than the two cheapest per-image baselines and at least ten times faster than the others, with less memory than any of them, since sampling a 36M U-Net never backpropagates through a surrogate.
These ratios hold even when the four images of a subject share the per-subject preparation.
The larger difference lies in how the cost grows over a collection.
A per-image optimizer spends $c_\text{id}\,N_\text{id} + c_\text{img}\,N_\text{img}$, where $c_\text{id}$ covers per-subject preparation such as the surrogate fine-tuning of Anti-DreamBooth and $c_\text{img}$ is a fresh optimization for every photograph.
PDM spends only $c_\text{img}^\text{PDM}\,N_\text{img}$, since it computes the target residual once and a single generator serves every subject.
A defender protecting an album therefore pays about one second per photograph, however many subjects the album contains.

Sampling with fewer steps makes PDM faster still (\fref{fig:steps}).
With 5 DDPM steps and no retraining, PDM protects an image in 0.16\,s, faster than RID, the fastest baseline, and still outperforms every baseline.
These five steps already keep most of the protection of the 50-step sample, so the number of sampling steps trades speed for protection at deployment, with one trained generator.

\begin{table}[t]
\centering
\caption{\textbf{Transfer under model mismatch.} Face protections are made on SD 1.5, and the personalization is retrained on SD 2.1 and SDXL, with FSS (lower is better).}
\label{tab:transfer}
\footnotesize
\setlength{\tabcolsep}{3pt}
\begin{tabular*}{\linewidth}{@{\extracolsep{\fill}}lccc@{}}
\toprule
Method & SD 1.5~\cite{rombach2022high} & $\to$ SD 2.1~\cite{rombach2022high} & $\to$ SDXL~\cite{podell2023sdxl} \\
\midrule
No protection & 0.406 & 0.454 & 0.662 \\
\midrule
PhotoGuard~\cite{salman2023raising} & 0.172 & 0.191 & 0.687 \\
ACE~\cite{zheng2023understanding} & 0.224 & 0.313 & 0.701 \\
Anti-DB~\cite{van2023anti} & 0.308 & 0.338 & 0.693 \\
AdvDM~\cite{liang2023adversarial} & 0.343 & 0.367 & 0.696 \\
CAAT~\cite{xu2024perturbing} & 0.334 & 0.364 & 0.685 \\
\textbf{PDM} & \textbf{0.070} & \textbf{0.076} & \textbf{0.461} \\
\bottomrule
\end{tabular*}
\end{table}

\smallskip
\noindent\textbf{Qualitative comparison.}
\fref{fig:qual} shows personalized generations under each protection.
Under the baselines overlaid noise and texture artifacts degrade the outputs, yet the subject usually stays recognizable behind them, in line with their intermediate identity scores in \tref{tab:comparison}.
Under PDM the generations often collapse into repetitive textures, the faces that do appear no longer resemble the subject, and on artworks the palette and brushwork of the artist disappear.
An output that degrades but keeps the subject recognizable still exposes the subject, whereas under PDM the personalized model reproduces little of the subject.

\smallskip
\noindent\textbf{Robustness to purification.}
In \tref{tab:purify} the attacker purifies the published images before fine-tuning on them, with five light purifications and two generative ones.
Under the light purifications, every baseline except MetaCloak~\cite{liu2024metacloak} loses most of its protection.
PDM, post-trained with the purification-aware reward (PAR), which scores each image after a random purification, keeps the strongest protection in every light-purification row on both domains, including the three purifications that it never sees in training.
Without PAR (PDM w/o PAR), the same generator keeps less under every light purification, so the robustness comes mostly from PAR.
Even so, PDM w/o PAR keeps more face protection than every baseline under every light purification, including MetaCloak, which optimizes its perturbation under random transformations (FSS 0.209 under JPEG against 0.338).
The two generative purifications regenerate the image with a large image model and remove most of the protection of every method~\cite{pleimling2026off,honig2025adversarial}.
Noisy upscaling is one of the strongest purifications in recent work~\cite{honig2025adversarial}, and no method keeps its protection under it.

\smallskip
\noindent\textbf{Transfer under model mismatch.}
\tref{tab:transfer} keeps the protected faces made on SD 1.5 and repeats the personalization on SD 2.1 and SDXL, which the perturbations were not optimized against, next to the SD 1.5 results on the same subjects.
On SD 2.1 PDM keeps nearly all of its protection and stays far ahead of every baseline, so its advantage does not depend on the victim being the surrogate.
SDXL replaces the autoencoder, the text encoders, and the backbone at once, so a perturbation shaped against the latent space of SD 1.5 no longer meets the features it was optimized for.
Even so, PDM still protects faces on SDXL, while every baseline stays at or above the unprotected level.

\begin{table}[t]
\centering
\caption{\textbf{Component ablation.} The upper block removes one component from the full PDM, and the lower block changes only how the perturbation is produced.}
\label{tab:ablation}
\footnotesize
\setlength{\tabcolsep}{2.5pt}
\begin{tabular*}{\linewidth}{@{\extracolsep{\fill}}lcccc@{}}
\toprule
& \multicolumn{2}{c}{Faces} & \multicolumn{2}{c}{Artworks} \\
\cmidrule(lr){2-3}\cmidrule(l){4-5}
Variant & FDFR $\uparrow$ & ISM $\downarrow$ & CSD $\downarrow$ & CLIP-SIM $\downarrow$ \\
\midrule
\textbf{Full (PDM)} & \textbf{0.412} & \textbf{0.118} & \textbf{0.145} & \textbf{0.668} \\
$-$ reward post-train (pre-only) & 0.206 & 0.171 & 0.270 & 0.724 \\
$-$ distillation pretrain (post-only) & 0.237 & 0.175 & 0.310 & 0.750 \\
$-$ purification-aware reward & 0.275 & 0.136 & 0.173 & 0.685 \\
\midrule
\textbf{PDM (50-step diffusion)} & \textbf{0.412} & \textbf{0.118} & \textbf{0.145} & \textbf{0.668} \\
One forward pass & 0.325 & 0.138 & 0.228 & 0.710 \\
Universal perturbation & 0.131 & 0.222 & 0.356 & 0.773 \\
\bottomrule
\end{tabular*}
\end{table}

\subsection{Model Analysis}
\label{sec:analysis}

\noindent\textbf{Component ablation.}
The upper block of \tref{tab:ablation} removes one component of PDM at a time, and every removal weakens protection.
Keeping only distillation (pre-only) or only reward post-training (post-only) falls well short of the full PDM.
The distilled model merely imitates the teacher and never sees the TRP loss directly, while reward post-training from a random initialization sees this loss but trains unstably and falls furthest on artworks.
Distillation gives post-training a stable start close to the teacher, and the reward then improves on it with the TRP loss itself, so only the two stages together reach the full protection.
Removing the purification-aware reward also weakens protection on images that the attacker does not purify.
We suspect that the random purifications act as input augmentation, so the perturbation stays effective on the victim whose weights move away from the frozen surrogate during fine-tuning.

The lower block of \tref{tab:ablation} changes only how the perturbation is produced, under the same TRP loss and data.
Amortizing the iterative teacher into one forward pass of the same U-Net loses protection on both domains ($-17\%$ in ISM and $-57\%$ in CSD relative to PDM), and this variant stays short of the PGD teacher, which reaches ISM 0.128 and CSD 0.157.
The many denoising steps of PDM give it the capacity to recover this loss and match the teacher on both domains.
A universal perturbation~\cite{moosavi2017universal}, optimized in the same way but fixed in advance and shared by every image, cannot adapt to the image it protects and loses far more ($-88\%$ and $-146\%$).
An amortized protector therefore has to produce the perturbation from each image and refine it over many steps, as PDM does.

\begin{table}[t]
\centering
\caption{\textbf{Budget response of the two geometries.} Exact matching (\eref{eq:mse}) and the projection of TRP (\eref{eq:trp}) under the same reference $r_t$ in the same PGD loop, with the residual norm $\|r_p\|$ that each attains and the protection after personalization.}
\label{tab:budget}
\footnotesize
\setlength{\tabcolsep}{3pt}
\begin{tabular*}{\linewidth}{@{\extracolsep{\fill}}lcccccc@{}}
\toprule
\multirow{2}{*}{Budget $\eta$} & \multicolumn{3}{c}{Exact matching} & \multicolumn{3}{c}{Projection (TRP)} \\
\cmidrule(lr){2-4}\cmidrule(l){5-7}
& $\|r_p\|$ & FSS $\downarrow$ & CSD $\downarrow$ & $\|r_p\|$ & FSS $\downarrow$ & CSD $\downarrow$ \\
\midrule
$2/255$ & 38.6 & 0.434 & 0.397 & 43.0 & 0.405 & 0.428 \\
$4/255$ & 36.9 & 0.464 & 0.393 & 46.0 & 0.338 & 0.398 \\
$8/255$ & 34.5 & 0.457 & 0.359 & 49.8 & 0.134 & 0.248 \\
$12/255$ & 33.2 & 0.360 & 0.315 & 52.1 & 0.069 & 0.157 \\
$16/255$ & 32.5 & 0.316 & 0.273 & 53.7 & 0.035 & 0.133 \\
\bottomrule
\end{tabular*}
\end{table}

\smallskip
\noindent\textbf{Objective analysis.}
We next analyze the two choices that define the TRP loss in \sref{sec:trp}, the geometry and the reference.
\tref{tab:budget} compares the two geometries under the same reference $r_t$, exact matching (\eref{eq:mse}) and the projection of TRP (\eref{eq:trp}).
Under exact matching the residual norm falls below its clean value of 40.5 at every budget, consistent with the magnitude term of \eref{eq:mse}, and its protection improves only slowly and on faces stays weaker than no protection up to $8/255$.
The projection grows the residual and turns each increase of the budget into additional protection.

\tref{tab:persist} pairs each reference ($q_{\mathrm{ACE}}$ or $r_t$) with each geometry (squared error or projection) in the same PGD loop.
It also includes the untargeted objective, which has no reference and raises only the magnitude term of \eref{eq:mse}.
The reference sets the direction.
With $q_{\mathrm{ACE}}$, the projection grows the residual about as much as TRP does, yet protection stays at the level of ACE, so a fixed direction alone is not enough and the target residual gives a better direction than the target latent.
The geometry sets the magnitude.
With $r_t$, exact matching keeps the alignment of TRP but shrinks the residual below that of the clean images, and on faces it protects less than ACE.
The untargeted objective grows the residual the most yet protects only weakly, so a large residual alone does not protect.
Only TRP, which pairs the target residual with the projection, protects strongly, so neither choice suffices alone.
This is consistent with the account of \sref{sec:trp}, in which a common target gives the errors of all protected images one shared pattern and one consistent bias~\cite{zheng2023understanding}, and only the projection grows this pattern without penalizing the size of the residual as the squared error does.

\begin{figure}[t]
\centering
\includegraphics[width=\linewidth]{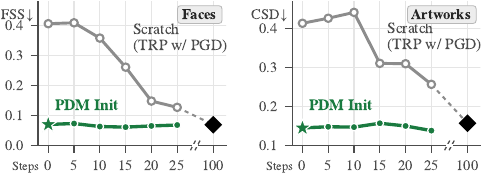}
\caption{\textbf{PDM sample against iterative optimization.} Protection after personalization (lower is better) over the steps of the TRP optimization, run from the clean image (Scratch) or from a PDM sample (PDM Init). The star at 0 steps is the PDM sample without any iteration. Scratch uses the step size of the PGD teacher, so the diamond at 100 steps is the teacher (TRP w/ PGD).}
\label{fig:refine}
\end{figure}

\smallskip
\noindent\textbf{Comparison with iterative optimization.}
\fref{fig:refine} compares a PDM sample with the TRP optimization that produces the PGD teacher.
Without any iteration, the sample protects about as strongly as the teacher after 100 steps and more strongly than 25 steps from the clean image, and more steps from the sample change protection little.
One PDM sample thus replaces the 100 steps of optimization that the teacher runs for each image.

\section{Conclusion}

We present PDM, a diffusion model that is trained once and protects a new image by sampling its perturbation, and the TRP loss, which grows the denoising residual along a target direction within the budget and supervises both training stages of PDM.
On faces and artworks, PDM outperforms nine baselines at the cost of one sampling pass and nearly matches its iterative teacher at a tenth of the teacher's cost.
At one second per image on consumer hardware, the cost no longer limits how many photographs a subject can protect.

\begin{table}[t]
\centering
\caption{\textbf{Reference and geometry of the objective.} Magnitude with no reference is the untargeted objective of AdvDM~\cite{liang2023adversarial}, and squared error with $q_{\mathrm{ACE}}$ is ACE~\cite{zheng2023understanding}. All rows run in the same PGD loop, so AdvDM and ACE differ from the other tables. $\|r_p\|$ is the residual norm on the protected faces before fine-tuning, and $\cos(r_p, r_t)$ is the cosine between $r_t$ and the residual of the fine-tuned victim on the protected images (faces / artworks).}
\label{tab:persist}
\footnotesize
\setlength{\tabcolsep}{3pt}
\begin{tabular*}{\linewidth}{@{\extracolsep{\fill}}llcccc@{}}
\toprule
Reference & Geometry & $\|r_p\|$ & $\cos(r_p, r_t)$ & FSS $\downarrow$ & CSD $\downarrow$ \\
\midrule
No protection & & 40.5 & $0.00$ / $0.00$ & 0.406 & 0.414 \\
\midrule
-- & Magnitude & 57.4 & $0.00$ / $0.00$ & 0.346 & 0.383 \\
$q_{\mathrm{ACE}}$ & Squared error & 48.5 & $+0.13$ / $+0.13$ & 0.255 & 0.311 \\
$q_{\mathrm{ACE}}$ & Projection & 51.4 & $+0.12$ / $+0.12$ & 0.247 & 0.307 \\
$r_t$ & Squared error & 33.2 & $+0.39$ / $+0.35$ & 0.360 & 0.315 \\
$r_t$ & Projection (ours) & 52.1 & $+0.37$ / $+0.34$ & \textbf{0.069} & \textbf{0.157} \\
\bottomrule
\end{tabular*}
\end{table}

\smallskip
\noindent\textbf{Limitations.}
Regenerating purifiers such as noisy upscaling remove most of the protection of every method in our evaluation~\cite{honig2025adversarial}.
Because PDM is trained once, its reward can include such purifiers and several surrogates with different backbones, which are too costly for per-image optimization, without adding cost at deployment.
Whether such training yields protection that survives regeneration and transfers to newer backbones remains open.

\section*{LLM Usage Statement}
LLMs were used to assist with code implementation and experiment management, and to edit the English prose of author-written drafts. All results come from executed experiments, and the authors reviewed and take full responsibility for all content.

{
    \small
    \bibliographystyle{ieeenat_fullname}
    \bibliography{main}

@inproceedings{rombach2022high,
  title={High-resolution image synthesis with latent diffusion models},
  author={Rombach, Robin and Blattmann, Andreas and Lorenz, Dominik and Esser, Patrick and Ommer, Bj{\"o}rn},
  booktitle={Proceedings of the IEEE/CVF Conference on Computer Vision and Pattern Recognition},
  year={2022}
}

@inproceedings{podell2023sdxl,
  title={SDXL: Improving latent diffusion models for high-resolution image synthesis},
  author={Podell, Dustin and English, Zion and Lacey, Kyle and Blattmann, Andreas and Dockhorn, Tim and M{\"u}ller, Jonas and Penna, Joe and Rombach, Robin},
  booktitle={International Conference on Learning Representations},
  year={2024}
}

@inproceedings{ruiz2023dreambooth,
  title={DreamBooth: Fine tuning text-to-image diffusion models for subject-driven generation},
  author={Ruiz, Nataniel and Li, Yuanzhen and Jampani, Varun and Pritch, Yael and Rubinstein, Michael and Aberman, Kfir},
  booktitle={Proceedings of the IEEE/CVF Conference on Computer Vision and Pattern Recognition},
  year={2023}
}

@inproceedings{gal2022image,
  title={An image is worth one word: Personalizing text-to-image generation using textual inversion},
  author={Gal, Rinon and Alaluf, Yuval and Atzmon, Yuval and Patashnik, Or and Bermano, Amit H and Chechik, Gal and Cohen-Or, Daniel},
  booktitle={International Conference on Learning Representations},
  year={2023}
}

@inproceedings{madry2017towards,
  title={Towards deep learning models resistant to adversarial attacks},
  author={Madry, Aleksander and Makelov, Aleksandar and Schmidt, Ludwig and Tsipras, Dimitris and Vladu, Adrian},
  booktitle={International Conference on Learning Representations},
  year={2018}
}

@article{ahn2024imperceptible,
  title={Imperceptible protection against style imitation from diffusion models},
  author={Ahn, Namhyuk and Ahn, Wonhyuk and Yoo, KiYoon and Kim, Daesik and Nam, Seung-Hun},
  journal={IEEE Transactions on Multimedia},
  volume={28},
  pages={4286--4300},
  year={2026}
}

@inproceedings{shan2023glaze,
  title={Glaze: Protecting artists from style mimicry by text-to-image models},
  author={Shan, Shawn and Cryan, Jenna and Wenger, Emily and Zheng, Haitao and Hanocka, Rana and Zhao, Ben Y},
  booktitle={USENIX Security Symposium},
  year={2023}
}

@inproceedings{van2023anti,
  title={Anti-DreamBooth: Protecting users from personalized text-to-image synthesis},
  author={Van Le, Thanh and Phung, Hao and Nguyen, Thuan Hoang and Dao, Quan and Tran, Ngoc N and Tran, Anh},
  booktitle={Proceedings of the IEEE/CVF International Conference on Computer Vision},
  year={2023}
}

@inproceedings{liang2023adversarial,
  title={Adversarial example does good: Preventing painting imitation from diffusion models via adversarial examples},
  author={Liang, Chumeng and Wu, Xiaoyu and Hua, Yang and Zhang, Jiaru and Xue, Yiming and Song, Tao and Xue, Zhengui and Ma, Ruhui and Guan, Haibing},
  booktitle={International Conference on Machine Learning},
  year={2023}
}

@article{liang2023mist,
  title={Mist: Towards improved adversarial examples for diffusion models},
  author={Liang, Chumeng and Wu, Xiaoyu},
  journal={arXiv preprint arXiv:2305.12683},
  year={2023}
}

@inproceedings{salman2023raising,
  title={Raising the cost of malicious AI-powered image editing},
  author={Salman, Hadi and Khaddaj, Alaa and Leclerc, Guillaume and Ilyas, Andrew and Madry, Aleksander},
  booktitle={International Conference on Machine Learning},
  year={2023}
}

@inproceedings{xue2023toward,
  title={Toward effective protection against diffusion-based mimicry through score distillation},
  author={Xue, Haotian and Liang, Chumeng and Wu, Xiaoyu and Chen, Yongxin},
  booktitle={International Conference on Learning Representations},
  year={2024}
}

@inproceedings{zheng2023understanding,
  title={Targeted Attack Improves Protection against Unauthorized Diffusion Customization},
  author={Zheng, Boyang and Liang, Chumeng and Wu, Xiaoyu},
  booktitle={International Conference on Learning Representations},
  year={2025}
}

@inproceedings{moosavi2017universal,
  title={Universal adversarial perturbations},
  author={Moosavi-Dezfooli, Seyed-Mohsen and Fawzi, Alhussein and Fawzi, Omar and Frossard, Pascal},
  booktitle={Proceedings of the IEEE Conference on Computer Vision and Pattern Recognition},
  year={2017}
}

@inproceedings{radford2021learning,
  title={Learning Transferable Visual Models From Natural Language Supervision},
  author={Radford, Alec and Kim, Jong Wook and Hallacy, Chris and Ramesh, Aditya and Goh, Gabriel and Agarwal, Sandhini and Sastry, Girish and Askell, Amanda and Mishkin, Pamela and Clark, Jack and Krueger, Gretchen and Sutskever, Ilya},
  booktitle={International Conference on Machine Learning},
  year={2021}
}

@article{heusel2017gans,
  title={GANs Trained by a Two Time-Scale Update Rule Converge to a Local Nash Equilibrium},
  author={Heusel, Martin and Ramsauer, Hubert and Unterthiner, Thomas and Nessler, Bernhard and Hochreiter, Sepp},
  journal={Advances in Neural Information Processing Systems},
  volume={30},
  year={2017}
}

@inproceedings{hu2021lora,
  title={LoRA: Low-Rank Adaptation of Large Language Models},
  author={Hu, Edward J and Shen, Yelong and Wallis, Phillip and Allen-Zhu, Zeyuan and Li, Yuanzhi and Wang, Shean and Wang, Lu and Chen, Weizhu},
  booktitle={International Conference on Learning Representations},
  year={2022}
}

@inproceedings{zhang2018unreasonable,
  title={The Unreasonable Effectiveness of Deep Features as a Perceptual Metric},
  author={Zhang, Richard and Isola, Phillip and Efros, Alexei A and Shechtman, Eli and Wang, Oliver},
  booktitle={Proceedings of the IEEE Conference on Computer Vision and Pattern Recognition},
  year={2018}
}

@article{saleh2015large,
  title={Large-scale classification of fine-art paintings: Learning the right metric on the right feature},
  author={Saleh, Babak and Elgammal, Ahmed},
  journal={International Journal for Digital Art History},
  number={2},
  year={2016}
}

@inproceedings{liu2024metacloak,
  title={MetaCloak: Preventing Unauthorized Subject-driven Text-to-image Diffusion-based Synthesis via Meta-learning},
  author={Liu, Yixin and Fan, Chenrui and Dai, Yutong and Chen, Xun and Zhou, Pan and Sun, Lichao},
  booktitle={Proceedings of the IEEE/CVF Conference on Computer Vision and Pattern Recognition},
  year={2024}
}

@inproceedings{xu2024perturbing,
  title={Perturbing Attention Gives You More Bang for the Buck: Subtle Imaging Perturbations That Efficiently Fool Customized Diffusion Models},
  author={Xu, Jingyao and Lu, Yuetong and Li, Yandong and Lu, Siyang and Wang, Dongdong and Wei, Xiang},
  booktitle={Proceedings of the IEEE/CVF Conference on Computer Vision and Pattern Recognition},
  year={2024}
}

@article{guo2024real,
  title={Real-time Identity Defenses against Malicious Personalization of Diffusion Models},
  author={Guo, Hanzhong and Nie, Shen and Du, Chao and Pang, Tianyu and Sun, Hao and Li, Chongxuan},
  journal={arXiv preprint arXiv:2412.09844},
  year={2024}
}

@inproceedings{poole2022dreamfusion,
  title={DreamFusion: Text-to-3D using 2D Diffusion},
  author={Poole, Ben and Jain, Ajay and Barron, Jonathan T and Mildenhall, Ben},
  booktitle={International Conference on Learning Representations},
  year={2023}
}

@inproceedings{clark2024directly,
  title={Directly Fine-Tuning Diffusion Models on Differentiable Rewards},
  author={Clark, Kevin and Vicol, Paul and Swersky, Kevin and Fleet, David J},
  booktitle={International Conference on Learning Representations},
  year={2024}
}

@inproceedings{karras2018progressive,
  title={Progressive Growing of {GAN}s for Improved Quality, Stability, and Variation},
  author={Karras, Tero and Aila, Timo and Laine, Samuli and Lehtinen, Jaakko},
  booktitle={International Conference on Learning Representations},
  year={2018}
}

@article{mittal2012no,
  title={No-Reference Image Quality Assessment in the Spatial Domain},
  author={Mittal, Anish and Moorthy, Anush Krishna and Bovik, Alan Conrad},
  journal={IEEE Transactions on Image Processing},
  volume={21},
  number={12},
  pages={4695--4708},
  year={2012}
}

@inproceedings{honig2025adversarial,
  title={Adversarial Perturbations Cannot Reliably Protect Artists From Generative {AI}},
  author={H{\"o}nig, Robert and Rando, Javier and Carlini, Nicholas and Tram{\`e}r, Florian},
  booktitle={International Conference on Learning Representations},
  year={2025}
}

@misc{vonplaten2022diffusers,
  title={Diffusers: State-of-the-art diffusion models},
  author={von Platen, Patrick and Patil, Suraj and Lozhkov, Anton and Cuenca, Pedro and Lambert, Nathan and Rasul, Kashif and Davaadorj, Mishig and Nair, Dhruv and Paul, Sayak and Berman, William and Xu, Yiyi and Liu, Steven and Wolf, Thomas},
  howpublished={GitHub repository},
  year={2022}
}

@inproceedings{pleimling2026off,
  title={Off-The-Shelf Image-to-Image Models Are All You Need To Defeat Image Protection Schemes},
  author={Pleimling, Xavier and Abdullah, Sifat Muhammad and Balde, Gunjan and Gao, Peng and Mondal, Mainack and Jadliwala, Murtuza and Viswanath, Bimal},
  booktitle={IEEE Conference on Secure and Trustworthy Machine Learning},
  year={2026}
}

@article{xu2023imagereward,
  title={ImageReward: Learning and evaluating human preferences for text-to-image generation},
  author={Xu, Jiazheng and Liu, Xiao and Wu, Yuchen and Tong, Yuxuan and Li, Qinkai and Ding, Ming and Tang, Jie and Dong, Yuxiao},
  journal={Advances in Neural Information Processing Systems},
  volume={36},
  year={2023}
}

@inproceedings{somepalli2024measuring,
  title={Investigating Style Similarity in Diffusion Models},
  author={Somepalli, Gowthami and Gupta, Anubhav and Gupta, Kamal and Palta, Shramay and Goldblum, Micah and Geiping, Jonas and Shrivastava, Abhinav and Goldstein, Tom},
  booktitle={European Conference on Computer Vision},
  year={2024}
}

@inproceedings{ahn2025nearly,
  title={Nearly Zero-Cost Protection Against Mimicry by Personalized Diffusion Models},
  author={Ahn, Namhyuk and Yoo, KiYoon and Ahn, Wonhyuk and Kim, Daesik and Nam, Seung-Hun},
  booktitle={Proceedings of the IEEE/CVF Conference on Computer Vision and Pattern Recognition},
  year={2025}
}

@inproceedings{umbach2024non,
  title={Non-Consensual Synthetic Intimate Imagery: Prevalence, Attitudes, and Knowledge in 10 Countries},
  author={Umbach, Rebecca and Henry, Nicola and Beard, Gemma and Berryessa, Colleen},
  booktitle={Proceedings of the CHI Conference on Human Factors in Computing Systems},
  year={2024}
}

@article{marchal2024generative,
  title={Generative AI Misuse: A Taxonomy of Tactics and Insights from Real-World Data},
  author={Marchal, Nahema and Xu, Rachel and Elasmar, Rasmi and Gabriel, Iason and Goldberg, Beth and Isaac, William},
  journal={arXiv preprint arXiv:2406.13843},
  year={2024}
}

@inproceedings{ronneberger2015u,
  title={U-Net: Convolutional Networks for Biomedical Image Segmentation},
  author={Ronneberger, Olaf and Fischer, Philipp and Brox, Thomas},
  booktitle={International Conference on Medical Image Computing and Computer-Assisted Intervention},
  year={2015}
}

@inproceedings{zhu2025gapdiff,
  title={{GAP-Diff}: Protecting {JPEG}-Compressed Images from Diffusion-based Facial Customization},
  author={Zhu, Haotian and Pang, Shuchao and Lu, Zhigang and Zhou, Yongbin and Xue, Minhui},
  booktitle={Network and Distributed System Security Symposium},
  year={2025}
}
}

\ifwithsupp
\clearpage
% \maketitlesupplementary
\appendix
\renewcommand{\thetable}{A\arabic{table}}
\renewcommand{\thefigure}{A\arabic{figure}}
\renewcommand{\theequation}{A\arabic{equation}}

    \section{Implementation Details}

    \subsection{Generator Architecture}

    PDM is a conditional U-Net trained from scratch with the $v$-prediction parameterization.
    The noised perturbation and the clean image form one six-channel input at $512\times512$, and the network predicts the velocity of the perturbation at the same resolution.
    The target residual $r_t$ is a four-channel SD~1.5 VAE latent.
    It is concatenated to the encoder features where they reach $64\times64$, at its native resolution, and a $1\times1$ convolution maps the result back to the encoder width; $r_t$ is neither upsampled to pixel space nor decoded.
    Self-attention appears once, at $32\times32$ in the middle block.
    \tref{tab:arch} lists the configuration.

    \subsection{Distillation Dataset}

    The teacher set holds 10{,}000 triples $(X_c, r_t, \delta^\star)$, each produced by one PGD run under the TRP loss at the shared budget $\eta = 12/255$.
    Clean images come from CelebA-HQ and WikiArt, 5{,}000 from each, and exclude every evaluation subject.
    One generator thus covers both domains without seeing the subjects it is evaluated on.
    Each image contributes a single perturbation.
    The generator gets its diversity from teacher settings that vary across the dataset.
    The teacher starts from $\delta = 0$, draws its step count uniformly from $\{50, \dots, 250\}$ and its step size from $\mathcal{U}[0.5, 2]/255$, and uses a fresh noise seed per example.

    All teacher runs, post-training, and deployment use one fixed target image.
    Its residual $r_t$ is computed once with the frozen SD~1.5 surrogate under an empty prompt, averaged over 20 timesteps spread across the full schedule with 5 noise samples each, and cached as a single tensor.

    Perturbations are stored as $\delta / \eta$ mapped to $[-1, 1]$ and quantized to 8-bit PNG, the same quantization a published image undergoes.
    The generator is therefore fit to perturbations that survive the format the defender actually distributes.
    The empty prompt is used for the teacher's PGD, for the cached residual, and at deployment; the conditioning never shifts between the three.

    \begin{table}[t]
    \centering
    \caption{\textbf{PDM generator configuration.} The generator operates on perturbations in pixel space and receives the target residual as a latent-resolution conditioning map.}
    \label{tab:arch}
    \footnotesize
    \begin{tabular*}{\linewidth}{@{\extracolsep{\fill}}ll@{}}
    \toprule
    Component & Setting \\
    \midrule
    Input & $\delta_s \oplus X_c$, 6 ch, $512\times512$ \\
    Output & velocity $v_\theta$, 3 ch, $512\times512$ \\
    Conditioning & $r_t$, 4 ch, concatenated at $64\times64$ \\
    Resolution ladder & $512\to256\to128\to64\to32$ \\
    Channels per resolution & 64, 128, 256, 512 \\
    Residual blocks & 1 per down stage, 1 per up stage \\
    Middle block & ResNet, self-attention, ResNet \\
    Attention resolution & $32\times32$ only \\
    Normalization & GroupNorm(32) + SiLU \\
    Timestep embedding & Sinusoidal \\
    Parameterization & $v$-prediction \\
    Noise schedule & Cosine, 1000 steps \\
    Parameters & 35.9M \\
    Deployment sampler & DDPM, 50 steps \\
    \bottomrule
    \end{tabular*}
    \end{table}

    \subsection{Training Hyperparameters}

    \tref{tab:hparam} summarizes the configurations of distillation and post-training.
    Post-training starts from the distilled checkpoint and uses the 50-step DDPM sampler of deployment.
    We exclude the first $30\%$ of the sampling trajectory, where noise still dominates the perturbation, and sample $K=2$ steps from the remaining trajectory.
    All unselected sampler steps run without gradients.
    At each selected step, we obtain the current perturbation estimate $\hat\delta_0$, project it onto the perturbation budget, and pass the resulting protected image through one purification drawn uniformly from no purification, JPEG compression with quality $\mathcal{U}[50, 90]$, bilinear down- and up-sampling with scale $\mathcal{U}[0.5, 0.75]$, and additive Gaussian noise with $\sigma \sim \mathcal{U}[0.01, 0.05]$ on the $[0, 1]$ scale.
    JPEG is applied as a real compression round trip and passes the gradient straight through.
    The TRP loss is then evaluated on the purified image.
    Each TRP-loss evaluation draws one random surrogate timestep and one noise sample.
    The losses from the selected steps are averaged, and the generator is updated once per post-training iteration.
    Post-training uses a constant learning rate and gradient clipping with a maximum norm of $1.0$, and runs for 5 epochs.
    The frozen SD~1.5 surrogate, the empty prompt, and the perturbation budget $\eta=12/255$ stay the same as in distillation.

    \begin{table}[t]
    \centering
    \caption{\textbf{Training configuration of the two stages.}}
    \label{tab:hparam}
    \footnotesize
    \begin{tabular*}{\linewidth}{@{\extracolsep{\fill}}lll@{}}
    \toprule
    & Distillation & Post-training \\
    \midrule
    Initialization & Random & Distilled checkpoint \\
    Optimizer & AdamW & AdamW \\
    Learning rate & $1\times10^{-4}$ & $1\times10^{-5}$ \\
    Weight decay & $1\times10^{-2}$ & $1\times10^{-2}$ \\
    LR schedule & Cosine annealing & Constant \\
    Batch size & 8 & 4 \\
    Training length & 100 ep. (125K upd.) & 5 ep. (12.5K upd.) \\
    Resolution & $512$ & $512$ \\
    Rollout steps & -- & 50 (DDPM) \\
    Selected steps $K$ & -- & 2 \\
    Purification in reward & -- & random, see text \\
    \bottomrule
    \end{tabular*}
    \end{table}

    \subsection{Training Cost}

    All training runs on RTX~5090 GPUs.
    The 10{,}000 teacher perturbations take 4.2\,h on four GPUs, about 17 GPU-hours, distillation takes 8.7\,h, and post-training takes 8.5\,h on a single GPU, about 34 GPU-hours in total.

    \subsection{Pseudo Code}

    Algorithm~\ref{alg:trp} evaluates the TRP loss that both training stages use.
    The target residual arrives cached, so the target branch costs no forward pass; the surrogate is evaluated once per timestep sample on the protected branch alone.
    Algorithm~\ref{alg:pre} generates the teacher data and fits the generator to it, and Algorithm~\ref{alg:post} optimizes the TRP loss in post-training.
    The algorithms write the perturbation in image units, as the main paper does, and the implementation runs the generator on $\delta/\eta \in [-1, 1]$.

    \begin{algorithm}[t]
    \caption{TRP loss}
    \label{alg:trp}
    \begin{algorithmic}[1]
    \REQUIRE image $X$, cached target residual $r_t$,
    frozen surrogate $\epsilon_\theta$, latent encoder $\mathcal{E}$,
    empty-prompt embedding $c$, samples $M$
    \STATE $z \leftarrow \mathcal{E}(X)$
    \STATE $\mathcal{L} \leftarrow 0$
    \FOR{$m=1$ \TO $M$}
        \STATE $\tau \sim \mathcal{U}\{0,\dots,T-1\}$,
        \quad $\epsilon \sim \mathcal{N}(0,I)$
        \STATE $z_\tau \leftarrow
        \alpha_\tau z+\sigma_\tau\epsilon$
        \STATE $r_p \leftarrow
        \epsilon_\theta(z_\tau,\tau,c)-\epsilon$
        \STATE $\mathcal{L} \leftarrow
        \mathcal{L}
        -\langle r_p,r_t\rangle/\|r_t\|_2$
    \ENDFOR
    \RETURN $\mathcal{L}/M$
    \end{algorithmic}
    \end{algorithm}

    \section{Detailed Experimental Setup}

    \subsection{Baseline Implementations}

    We run every per-image baseline based on its official implementation and change only the budget to $\eta = 12/255$ and the surrogate to SD~1.5.
    Step sizes below are on the $[-1, 1]$ image scale.
    PhotoGuard follows the texture-only mode of the Diff-Protect repository, pulling the latent of the protected image toward that of the repository's default target image under a latent squared error, with 100 steps of size $2/255$ from a random start.
    Mist adds to the same encoder term the diffusion training loss weighted by $10^4$, with the same steps and target.
    AdvDM ascends the diffusion training loss for 100 steps of size $2/255$ from a random start, and SDS uses the $(-)$ variant highlighted by its authors, with the same steps.
    ACE follows the Mist-v2 implementation, alternating 5 rounds of 30 PGD steps of size $5\times10^{-3}$ toward the same target with 10 LoRA fine-tuning steps of rank 4 on the subject images.
    Anti-DreamBooth runs the ASPL loop of 50 rounds, each fine-tuning a copy of the full U-Net and text encoder for 3 steps at learning rate $5\times10^{-7}$ and then taking 6 PGD steps of size $5\times10^{-3}$; its official scripts use SD~2.1 as the surrogate, which we replace with SD~1.5 to match the victim.
    CAAT runs 250 steps of size $5\times10^{-3}$ while fine-tuning the cross-attention keys and values at learning rate $10^{-5}$.
    MetaCloak keeps its pool of surrogate models and its random transformations.
    RID uses the released network trained at $\eta = 12/255$ with its official preprocessing, which matches ours.
    The PGD teacher of PDM runs 100 TRP steps of size $1/255$ from $\delta = 0$ with the same target residual as PDM.

    \subsection{Victim Personalization and Metrics}

    \noindent\textbf{Personalization.} The victim trains a rank-4 LoRA adapter on SD~1.5 for 300 steps at learning rate $1\times10^{-4}$ with the official \texttt{diffusers} script~\cite{vonplaten2022diffusers}, with the text encoder frozen and without prior preservation, on the four protected images of one subject, using the prompt ``a photo of sks person'' for faces and ``a painting of sks style'' for artworks.
    Generation uses 50 Euler ancestral steps at guidance scale 7, producing 16 images per subject from sampling seeds shared across methods, and every protection is evaluated through this identical chain.

    \begin{algorithm}[t]
    \caption{Teacher-data generation and distillation pre-training}
    \label{alg:pre}
    \begin{algorithmic}[1]
    \REQUIRE clean-image pool $\mathcal{X}_c$,
    cached target residual $r_t$, budget $\eta$
    \STATE initialize an empty dataset $\mathcal{D}$
    
    \STATE \textbf{// teacher-data generation}
    \FOR{each training example}
        \STATE $X_c \sim \mathcal{X}_c$
        \STATE $N \sim \mathcal{U}\{50,\dots,250\}$
        \STATE $\gamma \sim \mathcal{U}[0.5,2]/255$
        \STATE $\delta \leftarrow 0$

        \FOR{$n=1$ \TO $N$}
            \STATE $g \leftarrow
            \nabla_\delta
            \mathcal{L}_{\mathrm{TRP}}(X_c+\delta,r_t)$
            \STATE $\delta \leftarrow
            \Pi_{\eta}
            \left(
            \delta-\gamma\,\mathrm{sign}(g)
            \right)$
        \ENDFOR
    
        \STATE $\delta^\star \leftarrow \delta$
        \STATE store $(X_c,r_t,\delta^\star)$ in $\mathcal{D}$
    \ENDFOR
    
    \STATE \textbf{// distillation pre-training}
    \REPEAT
        \STATE $(X_c,r_t,\delta^\star)\sim\mathcal{D}$
        \STATE $s\sim\mathcal{U}\{0,\dots,T-1\}$
        \STATE $\epsilon\sim\mathcal{N}(0,I)$
        \STATE $\delta_s\leftarrow
        \alpha_s\delta^\star+\sigma_s\epsilon$
        \STATE $v_s\leftarrow
        \alpha_s\epsilon-\sigma_s\delta^\star$
        \STATE $\mathcal{L}_{\mathrm{distil}}\leftarrow
        \left\|
        v_s-v_\theta(\delta_s,s,X_c,r_t)
        \right\|_2^2$
        \STATE update $\theta$ using
        $\nabla_\theta\mathcal{L}_{\mathrm{distil}}$
    \UNTIL{converged}
    \end{algorithmic}
    \end{algorithm}

    \begin{algorithm}[t]
    \caption{Post-training}
    \label{alg:post}
    \begin{algorithmic}[1]
    \REQUIRE distilled $\theta$,
    DDPM timestep schedule $\mathcal{T}$ of length $S=50$,
    selected steps $K=2$, skip fraction $\rho=0.3$,
    budget $\eta$, purification set $\mathcal{P}$
    
    \REPEAT
        \STATE $(X_c,r_t)\sim\mathcal{D}$
        \STATE $\delta\sim\mathcal{N}(0,I)$
        \STATE $\mathcal{I}_{\mathrm{valid}}
        \leftarrow
        \{\lfloor\rho S\rfloor+1,\dots,S\}$
        \STATE draw $K$ distinct indices
        $\mathcal{I}_{\mathrm{sel}}$
        uniformly from $\mathcal{I}_{\mathrm{valid}}$
        \STATE $\mathcal{L}_{\mathrm{post}}\leftarrow0$
    
        \FOR{$i=1$ \TO $S$}
            \STATE $s\leftarrow\mathcal{T}[i]$
    
            \IF{$i\in\mathcal{I}_{\mathrm{sel}}$}
                \STATE $v\leftarrow
                v_\theta(\delta,s,X_c,r_t)$
                \quad // with gradient
                \STATE $\hat{\delta}_0\leftarrow
                \mathrm{to\_}\delta_0(v,\delta,s)$
                \STATE $\tilde{\delta}\leftarrow
                \Pi_{\eta}
                (\hat{\delta}_0)$
                \STATE $\mathrm{pur}\sim\mathcal{P}$
                \quad // one random purification
                \STATE $\mathcal{L}_{\mathrm{post}}
                \mathrel{+}=
                \mathcal{L}_{\mathrm{TRP}}
                (\mathrm{pur}(X_c+\tilde{\delta}),r_t;\,M=1)$
            \ELSE
                \STATE $v\leftarrow
                v_\theta(\delta,s,X_c,r_t)$
                \quad // without gradient
            \ENDIF
    
            \STATE $\delta\leftarrow
            \operatorname{stopgrad}
            \bigl(\mathrm{step}(v,\delta,s)\bigr)$
        \ENDFOR
    
        \STATE $\mathcal{L}_{\mathrm{post}}
        \leftarrow\mathcal{L}_{\mathrm{post}}/K$
        \STATE update $\theta$ using
        $\nabla_\theta\mathcal{L}_{\mathrm{post}}$,
        clip gradient norm at $1.0$
    \UNTIL{converged}
    \end{algorithmic}
    \end{algorithm}

\smallskip
    \noindent\textbf{Metrics.} FDFR counts the generations in which RetinaFace detects no face at confidence threshold 0.9 on the generation padded by 25\%.
    Generations with a detected face but no extracted embedding are excluded from FDFR, so FDFR counts detection failures only.
    ISM is the mean cosine similarity between the ArcFace embedding of each detected face and the mean embedding of the subject's clean images.
    FSS combines the two per subject as $(1 - \mathrm{FDFR}) \times \mathrm{ISM}$, counting a subject with no detected generation as zero, and is then averaged over subjects.
    CSD uses the released ViT-L style descriptor of~\cite{somepalli2024measuring} and CLIP-SIM uses CLIP ViT-B/32, each as the mean cosine between generations and the subject's clean images.
    BRISQUE uses the \texttt{pyiqa} implementation.
    FID uses \texttt{clean-fid} in its \texttt{clean} mode, pooled over all subjects of one method and domain (320 generations), since the 16 generations of one subject are too few for its 2048-dimensional statistic.
    Its reference is the clean images of the same domain in the teacher set, about 5{,}000 per domain that exclude every evaluation subject, identical across methods.

    The main comparison and the sampling-step sweep use all 20 subjects of each domain, and the other experiments use the first 10 subjects.

    \subsection{Cost of Protecting an Image}
    \label{sec:supp-cost}

    Timings are wall-clock on a single RTX~5090 at $512\times512$, with every method processing one image at a time.
    For each method we time the whole process for one and for four images of a subject, three runs each, and fit $T(N) = L + cN$ to the medians; the slope $c$ is the per-image cost and the intercept $L$ is model loading, about 12\,s for the methods that instantiate the SD~1.5 surrogate and 9.5\,s for PDM.
    For ACE, Anti-DreamBooth, CAAT, and MetaCloak, which prepare a surrogate per subject, the intercept exceeds the loading time, and the excess is their per-subject term (15.7, 60.6, 21.5, and 1081.7\,s), which \tref{tab:comparison} of the main paper adds to the per-image cost.
    TRP optimized with PGD takes 10.40\,s and 5.11\,GB for 100 steps.
    Peak memory is the maximum that \texttt{nvidia-smi} reports when polled every 200\,ms.
    MetaCloak is omitted from \fref{fig:steps} of the main paper because its cost of 1{,}667\,s lies outside the plotted range.

    \subsection{Purification}

    Five purifications require no learning, namely JPEG at quality 75, bilinear downsampling to $0.5\times$ followed by bicubic upsampling back to $512$, WebP at quality 75, Gaussian blur with $\sigma = 1$, and Lanczos downsampling to $0.5\times$ followed by Lanczos upsampling.
    Post-training does not use WebP, blur, or Lanczos resizing.
    Two generative purifications are taken from recent work, and for both we follow the authors' own settings.
    Noisy upscaling~\cite{honig2025adversarial} adds Gaussian noise of $\sigma = 0.1$ on the $[0, 1]$ scale and passes the image through the SD $\times4$ upscaler at noise level 320 before Lanczos resizing back to $512$.
    FLUX img2img~\cite{pleimling2026off} passes the image through FLUX.1-dev at strength 0.15 with the prompt ``Denoise the image'' and the pipeline's default resolution, steps, and guidance, before Lanczos resizing back to $512$.
    Purified copies then enter the same personalization and evaluation chain as the unpurified ones, so a row differs from the None row only in the purification.

    \subsection{Transfer under Model Mismatch}

    Protections are made once on SD~1.5 and reused unchanged, and only the base model of the victim changes, to SD~2.1-base at $512$ and to SDXL-base-1.0 at $1024$ with the fp16-corrected autoencoder.
    Both victims use the identifier token \texttt{ohwx}, since \texttt{sks} has a strong rifle prior in their larger text encoders and keeps the subject from attaching to the token.
    SD~2.1-base otherwise keeps the LoRA rank, learning rate, and 300 steps of the SD~1.5 recipe.
    SDXL follows the official \texttt{diffusers} DreamBooth LoRA recipe for SDXL, with 500 steps at learning rate $10^{-4}$ in fp16 with batch size 1 and 4 gradient-accumulation steps, sampled with its default Euler sampler at guidance scale 5.
    The training script resizes the $512\times512$ protected images to $1024$ with Lanczos interpolation for every method.

    \section{Additional Qualitative Results}

    \fref{fig:qual-sup} repeats the comparison of \fref{fig:qual} in the main paper on additional face and artwork subjects.

    \begin{figure*}[t]
    \centering
    \includegraphics[width=\textwidth]{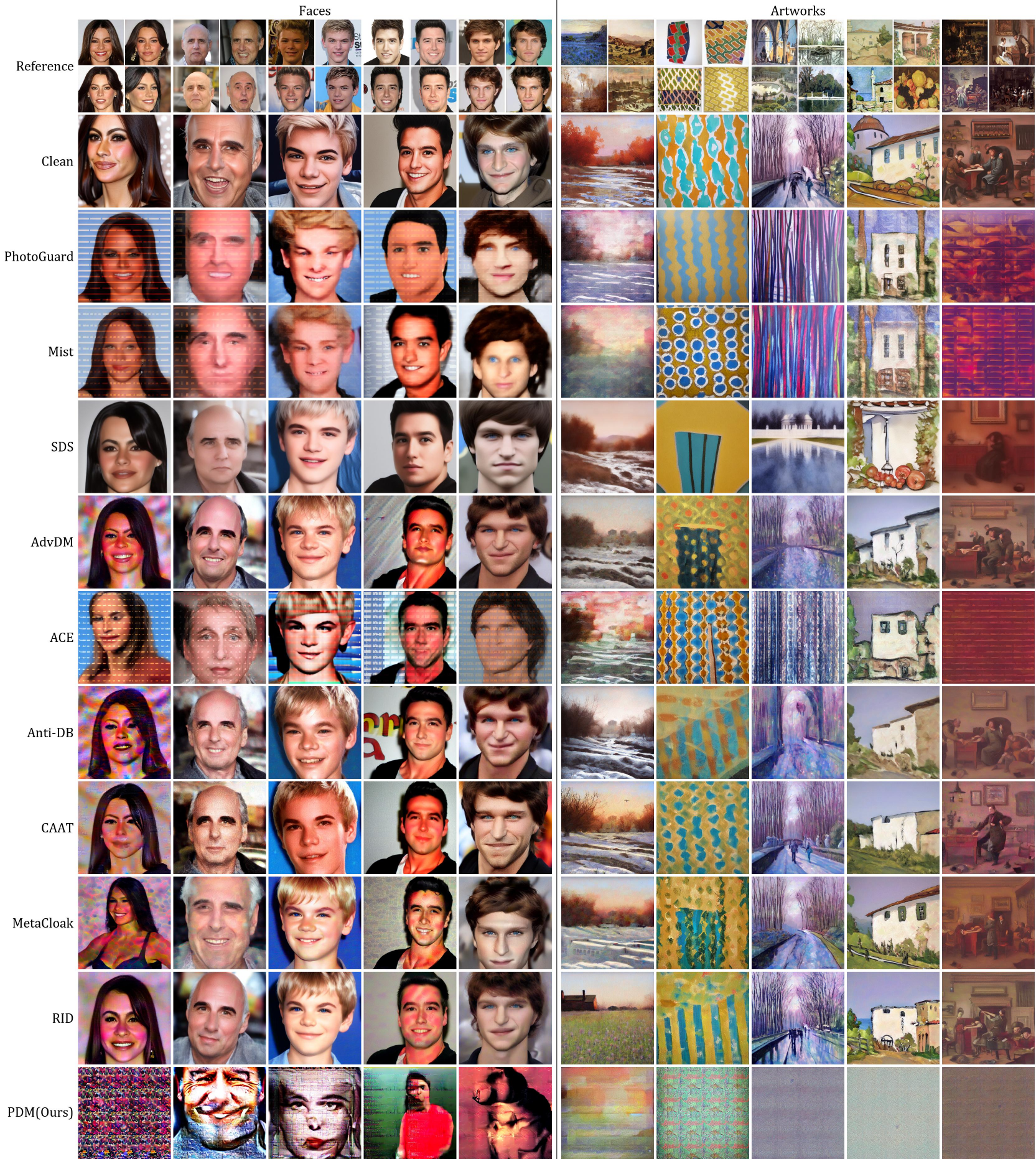}
    \caption{\textbf{Additional qualitative comparison.} The top row shows the training images of each subject and the remaining rows show personalized generations under each protection. Under the baselines the subject often stays recognizable behind the added noise and texture, whereas under PDM the generations often collapse into repetitive texture and the faces that appear no longer resemble the subject.}
    \label{fig:qual-sup}
    \end{figure*}

\fi

\end{document}